\documentclass[11pt,a4paper]{article}

\usepackage{graphicx}
\usepackage{booktabs}
\usepackage{array}
\usepackage{multirow}
\usepackage{float}
\usepackage{amsmath}
\usepackage{hyperref}
\usepackage{subfigure}
\usepackage{natbib}
\usepackage{tikz}
\usepackage{units}
\usepackage{lineno}
\usepackage[margin=2.5cm]{geometry}

\usepackage{xcolor}
\begin{document}

\title{Agent-Based Post-Hoc Correction of Agricultural Yield Forecasts}
\author{Matthew Beddows\\
School of Natural and Computing Sciences\\
University of Aberdeen, UK\\
\texttt{m.beddows.21@abdn.ac.uk}
\and
Aiden Durrant\\
School of Computing Sciences\\
University of East Anglia, UK\\
\texttt{aiden.durrant@uea.ac.uk}
\and
Georgios Leontidis\\
Department of Physics and Technology \\
UiT The Arctic University of Norway\\
\texttt{georgios.leontidis@uit.no}
}

\maketitle

\begin{abstract}
Accurate crop yield forecasting in commercial soft fruit production is constrained by the data available in typical commercial farm records, which lack the sensor networks, satellite imagery, and high-resolution meteorological inputs that most state-of-the-art approaches assume. We propose a structured LLM agent framework that performs post-hoc correction of existing model predictions, encoding agricultural domain knowledge across tools for phase detection, bias learning, and range validation. Evaluated on a proprietary strawberry yield dataset and a public USDA corn harvest dataset, agent refinement of XGBoost reduced MAE by 20\% and MASE by 56\% on strawberry, with consistent improvements across Moirai2 (MAE$-$24\%, MASE$-$22\%) and Random Forest (MAE$-$28\%, MASE$-$66\%) baselines. Using Llama 3.1 8B as the agent produced the strongest corrections across all configurations; LLaVA 13B showed inconsistent gains, highlighting sensitivity to the choice of refinement model.

\end{abstract}


 \section{Introduction}
 Accurate yield forecasting is essential for labour allocation, cold storage, and retail contract fulfilment for UK strawberry producers \cite{celis2024review}, yet most state-of-the-art approaches assume sensor networks, satellite imagery, or high-resolution meteorological records that commercial growers simply do not access to The UK soft fruit industry is further constrained by the fact that 95\% of strawberries are grown under polytunnels \cite{batke2025protected}, rendering satellite and UAV-based collection impractical regardless of cost. In practice, a typical farm record consists of a few years of weekly yield figures, basic plot metadata, and externally sourced weather data.This creates a persistent gap between what the literature demonstrates and what can realistically be deployed.

One way to address this is to acquire additional data, but this can often be  difficult for growers due to both monetary and time costs, and this would also rule out years of previous historical data they cannot retroactively improve. Our method takes a different approach, reasoning more effectively over the data that is already available.

The core problem we address is that existing forecasting models even when 
provided with recent historical observations as an input, continue to exhibit 
systematic errors, that growers can easily distinguish as errors. We propose that explicitly encoding this reasoning into a structured agent layer produces consistent improvements over the base model, without requiring additional data or retraining.

\begin{figure*}[h]
    \centering
     \includegraphics[width=1.0\textwidth]{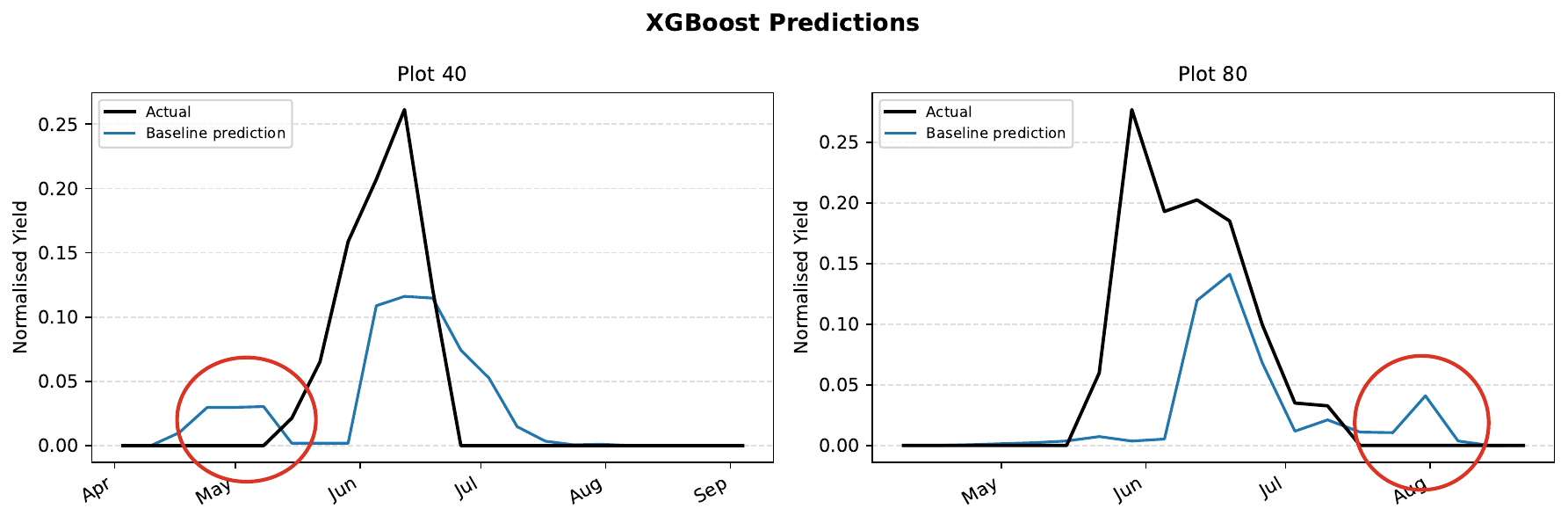}
     \caption{XGBoost research baseline predictions on two test plots, showing a pre-season spike (left) and a post-season spike (right). Both illustrate how a model trained on windowed numerical features, without explicit seasonal constraints, can produce predictions that contradict the crop's growth cycle.}
     \label{fig:model_xgb}
 \end{figure*}

Figure~\ref{fig:model_xgb} shows two characteristic failure modes. In both cases a grower would immediately reject the forecast: strawberries do not yield before flowering, and fields do not recover after clearance. 
The model has no mechanism to enforce either constraint. The model sees only numbers, it has no representation of where the crop is in its growing cycle.

This is the gap the agent fills. The base model's prediction is taken as-is; the agent then checks it against the crop's current seasonal phase, recent error history, and the training distribution, and corrects it where the prediction contradicts known seasonal behaviour. 

\begin{figure}[h]
     \centering
     \includegraphics[width=\linewidth]{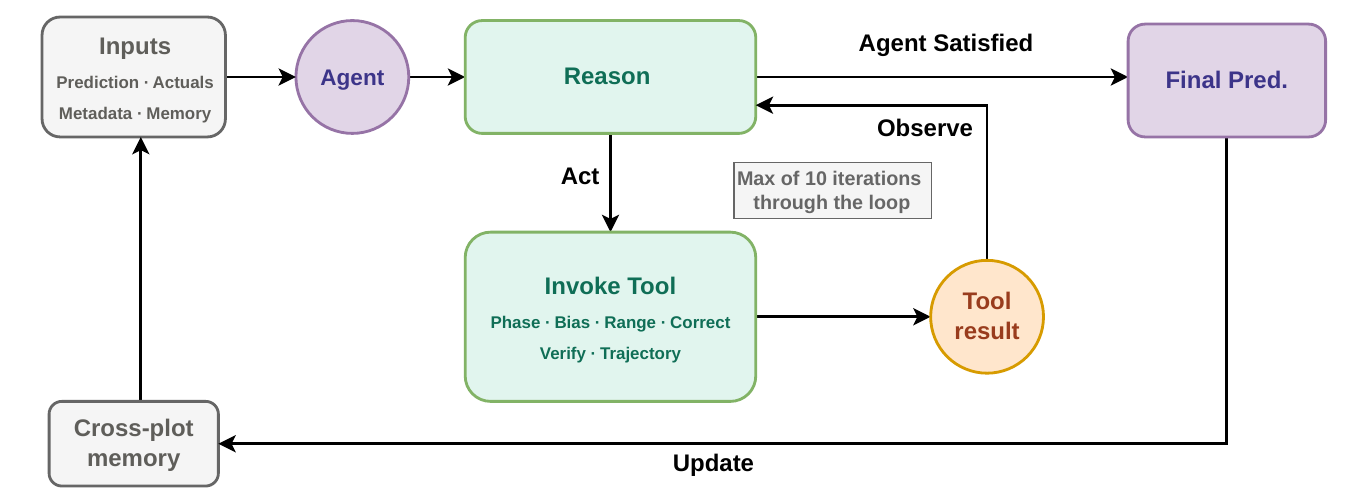}
    \caption{Overview of the agent pipeline. The ReAct loop iterates over the tool library to refine the raw prediction, the accepted correction is passed to the cross-plot memory, which is injected into the agent prompt on subsequent predictions.}
     \label{fig:system-overview}
 \end{figure}

This is the gap the agent fills. The base model's prediction is taken as-is; the agent then checks it against the crop's current seasonal phase, recent error history, and the training distribution, and corrects it where the prediction contradicts known seasonal behaviour. Figure~\ref{fig:system-overview} shows the overall system architecture.

\subsection{Novelty and Contributions}
Our research demonstrates that an agent providing LLM-based post-hoc prediction correction can consistently improve yield forecasts for agricultural applications.

This work makes the following contributions:

 \begin{itemize}
    \item We propose a structured LLM agent framework for \textit{post-hoc 
           prediction correction}, operating on top of an existing forecasting 
           model rather than replacing it — a largely unexplored field in 
          time series forecasting.
          
    \item The agent encodes agricultural domain knowledge into a fixed toolset, 
    covering seasonal phase detection, bias learning, range validation, and 
    correction synthesis, where each correction can be traced to a specific, 
    inspectable reasoning step.

    \item The method is evaluated on two seasonal datasets of different crops from          different origins.The datasets include a proprietary polytunnel strawberry        dataset and the public USDA corn harvest dataset. The method achieves MAE         reductions of up to 28\% on strawberry and 24\% on corn across three              baseline forecasting models, demonstrating that the approach generalises          across domains and models.
    
 \end{itemize}

\section{Related Work}

\subsection{Machine Learning in Agrifood}

Globally, there has been substantial advancements in the integration of AI into agricultural practices, particularly in precision farming, crop yield forecasting, and logistics/resource management \cite{arevalo2025ai,oliveira2023artificial}. However, even with all these developments, there are still many limitations that remain around data availability, system interoperability, model development, deployment, and adoption {\cite{durrant2022role,chen2024empowering,cricelli2024technological}.

Machine learning methods have already proven effective in many agricultural sectors \cite{alhnaity2019using,uppalapati2025precision,suarez2024forecasting}. In particular, some studies have highlighted the accuracy of these methods for predicting strawberry yields \cite{onoufriou2023premonition,liu2025ai}.

Although these studies reported positive outcomes with transformer models and extensive datasets \cite{onoufriou2023premonition,lee2020framework}, they depended on high-resolution information from irrigation infrastructure, meteorological stations, and high resolution harvest records. Often crop yield forecasting relies on extensive meteorological data, remote sensing information, and soil characteristics \cite{xiao2025progress}, which are not available to many growers.

Often approaches have utilized satellite imagery and soil parameters \cite{chaudhary2021deep} or UAV-mounted cameras \cite{zheng2022prediction}. These methods have shown promise, but the UK Soft Fruit industry is very reliant on polytunnels, with 95\% of UK strawberries being grown within them \cite{batke2025protected}; this makes these solutions impractical. More modern techniques such as transformer-based time-series models have also been successful in crop yield forecasting. However, once again, these models rely on comprehensive and/or multi-modal datasets \cite{lin2023mmst, onoufriou2023premonition}. In our prior work, foundation time series models were combined with traditional ML techniques to navigate this constraint \cite{beddows2025visiontrees}; however this is limited to univariate data, which excludes a lot of important data growers routinely collect on their crops. This motivated our alternative approach: rather than requiring additional data modalities as a prerequisite, the system instead reasons more effectively over the inputs already available to the forecasting pipeline.

\subsection{LLM Agents for Structured Tasks}
The rise of LLM agents has changed how we automate complex tasks by moving away from static prompts toward iterative loops of reasoning and tool use. This "ReAct" style approach \cite{yao2022react} allows a model to plan a step, execute an external query, and then adjust its logic based on the result. In time-series work, this has mostly involved using LLMs as sophisticated feature extractors or using fine-tuned models to interpret data directly \cite{jin2024tsllmsurvey}. Models like Time-LLM \cite{jin2023time}, for instance, treat time-series data as a linguistic pattern-matching problem to generate forecasts. This trend extends into finance, where researchers are combining historical prices with news sentiment to build more explainable, multi-modal forecasting tools \cite{yu2023harnessing}. Some have even used LLMs to "guess" the underlying structure of a dataset to help train Graph Neural Networks, often beating traditional deep learning benchmarks \cite{chen2023chatgpt}.

However, there is a skepticism regarding whether LLMs are actually "reasoning" or just executing complex pattern recognition. In many structured tasks, standard LLMs still struggle to beat small, specialized models that are purpose-built for relational logic \cite{li2024llms}. Even if we set this "reasoning" debate aside, some models are still proving useful, for example, the DCATS framework \cite{yeh2025empowering} uses an agent to clean and refine training data, which reportedly cuts error rates by 6\% across various models. In the agricultural domain specifically, Mandiga et al.\ \cite{mandiga2026nutrichat} demonstrated that a ReAct-based agent with expert-designed tools substantially outperforms standalone LLMs on structured poultry nutrition queries, with tool-grounded responses reducing hallucination by 62\% and improving numerical precision by over 160\% compared to GPT-4o. Their work highlights that encoding domain knowledge directly into a fixed toolset, rather than relying on the LLM's parametric knowledge alone, is a meaningful design decision that determines whether an agent produces reliable outputs in specialist settings.

The current gap in the literature is that LLMs are almost always used before or during the forecasting process, for data prep or for the prediction itself. NutriCHAT \cite{mandiga2026nutrichat} and similar systems ground the agent in domain knowledge at query time, but the prediction itself remains unchallenged. Very little work has been done on post-hoc correction, or using an agent to fix a prediction that has already been made. In many industries, the "base" forecasting models are already decent, but they often make systematic errors that they lack the context to fix. We propose an agent-based layer that sits on top of these existing predictors. Instead of asking the LLM to predict the yield, we use it to apply seasonal and statistical logic to the model's output, using the same messy data growers rely on: yield history, plot metadata, and weather patterns.

\section{Datasets}
\label{sec:datasets}
We evaluate our method on two seasonal datasets. A proprietary soft fruit yield dataset collected in partnership with a commercial grower Angus Soft Fruits, and a crop harvest dataset created from publicly available data from the USDA, chosen to demonstrate that the method is not dependent on domain-specific data infrastructure.

\subsection{Angus Soft Fruits}
To demonstrate real-world applicability, we collaborated with Angus Soft Fruits Ltd, a large commercial supplier of soft fruit to UK and European retailers. The dataset comprises strawberry yield records from 2020 onward, covering 123 plots across 16 farms in Scotland and England. Variables include date, yield, farm and plot identifiers, plot acreage, variety, and tunnel type. Data from 2020--2022 form the training set and 2023 the test year, with 80 plots used for evaluation.

Yield records were supplemented with ERA5-Land reanalysis weather data matched to each farm's location and week, including temperature, dewpoint temperature, total precipitation, surface pressure, and wind components aggregated to a daily resolution. ERA5-Land was used in place of on-site weather observations, which 
were unavailable, and has been validated as a reliable substitute in similar agricultural contexts\cite{beddows2024multi}.

\subsection{USDA Corn Harvest}
To validate the method on a second seasonal dataset which could be publicly accessed, weekly corn data was sourced from the USDA National Agricultural Statistics Service (NASS) QuickStats database across 38 US states from 2020 to 2023, with 2020-2022 used for training and 2023 held out as the test year. Three data items were retrieved, harvest progress (PCT HARVESTED), used as the primary yield signal, crop condition (PCT EXCELLENT and PCT GOOD), combined into a seasonal health indicator and planting progress (PCT PLANTED), used to characterise whether each state's season started early or late relative to historical norms. USDA reports harvest progress as a cumulative weekly percentage, consecutive values were differenced to recover a week-by-week harvest rate. The resulting series follows the same bell-curve structure as the strawberry data, rising from zero at the start of the season, peaking mid-autumn, and returning to zero once harvest is complete.

\subsection{Feature engineering}
A 2-week prediction horizon was applied, meaning the most recent yield observation available to the model at prediction time is from two weeks prior. For XGBoost and Random Forest, the input window covers weeks $t{-}5$ to $t{-}2$, creating a deliberate gap that reflects the real-world constraint of knowing only what has already been harvested. For the strawberry dataset ERA5 weather features were windowed identically. Categorical variables (tunnel type, farm and plot identifiers) were one-hot encoded. Rather than training on all available plots, the system selects the top $K=50$ most similar historical plots by DTW distance as the training context for each test entity.

\section{Methodology}
The proposed system is a hybrid framework combining conventional ML forecasting with a Task-Oriented Autonomous Agent using a ReAct (Reason + Act) orchestration loop. Rather than building a more complex forecasting model, we add a reasoning layer on top of an existing one. A conventional ML model generates the initial prediction, and the agent then examines that prediction in context with recent yield history, the crop's current growth stage, systematic errors the model has been making, similar historical plots retrieved from a knowledge graph and data-driven guardrails derived from the training distribution to decide whether and how to correct it.

\subsection{System Pipeline Overview}
Figure~\ref{fig:model_visionTS} gives an overview of the full pipeline. The system operates in two phases: knowledge graph construction from historical training plots, followed by a week-by-week prediction loop in which the agent corrects each forecast in turn.

\begin{figure}[h]
    \centering
    \includegraphics[width=\linewidth]{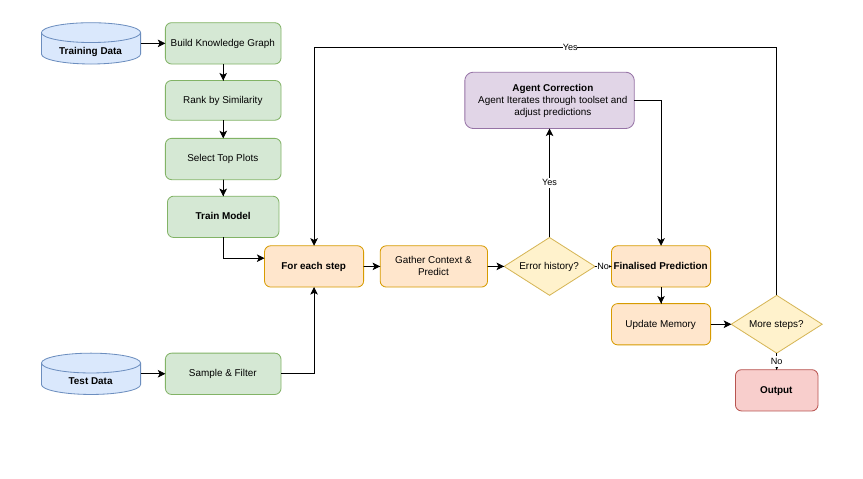}
    \caption{High-level pipeline overview. Training data is encoded into the knowledge graph; test data is prepared and fed into the weekly prediction loop one step at a time.}  
    \label{fig:model_visionTS}
\end{figure}

\subsubsection{Phase 1: Knowledge Graph Construction}
A grower's understanding of their operation is built up over many seasons, not reconstructed from scratch each year. The knowledge graph mirrors this by encoding the character of every training plot as a structured set of curve shape features, capturing not just raw yield figures but the geometric and seasonal properties that define how each plot behaves.

\subsubsection{Phase 2: Prediction Loop}
As the season progresses, a grower's assessment of one plot informs how they read any other. The prediction loop replicates this by advancing all test plots one week at a time, allowing the agent to learn from plots already processed while never drawing on observations from future weeks. This lockstep structure ensures the agent's accumulated knowledge reflects only what would genuinely be available at each point in the season. The agent's reasoning loop, toolset, and correction pipeline are described in full in Section~\ref{sec:prediction_loop}.

\begin{table*}[h]
\centering
\caption{Contents of the agent state vector $S$ at each prediction step.}
\label{tab:state_vector}
\begin{tabular}{ll}
\toprule
\textbf{Component} & \textbf{Used by} \\
\midrule
Raw prediction and prediction interval  & All tools \\
Recent actuals                          & Bias Learning, Phase Detection, Range Validation \\
Recent predictions                      & Bias Learning \\
Lookahead prediction                    & Phase Detection \\
Forecast horizon                        & Phase Detection \\
Seasonal position                       & Apply Correction, Position bias table \\
Known lagged error                      & Bias Learning, Position bias table \\
Entity metadata                         & Similarity Search, Agent prompt \\
Historical plot data                    & Range Validation, Safety Verifier \\
Jump distribution                       & Safety Verifier \\
\bottomrule
\end{tabular}
\end{table*}

\begin{itemize}
    \item \textbf{Observation:} At each correction step the agent receives the     state vector $S$ summarised in Table~\ref{tab:state_vector}. The \textbf{Jump Distribution} is the empirical distribution of consecutive week-on-week ratios $y_t / y_{t-1}$ computed across all training plots, and summarised as $P_{01}$--$P_{99}$ percentiles is included as a data-driven guardrail for \textbf{Safety Verifier}.
    
    \item \textbf{Decision Logic:} The agent selects tools based on the current state via a ReAct loop of up to 10 iterations. At each iteration it receives a prompt containing the current state and responds with a reason and a tool to call next. The loop exits early when the agent returns no further tools.
    The \textbf{Safety Verifier} is forced automatically after every correction call.
\end{itemize}

\subsection{Knowledge Graph Construction}
\label{sec:kg_construction}
A grower's accumulated knowledge of their plots is not stored as raw yield figures but as an intuitive sense of each crop's character: how volatile it is, when it typically peaks, how quickly it declines. The knowledge graph formalises this by encoding each training entity as a node carrying two types of attributes: metadata from the data loader (farm, variety, plot size, etc.) and a set of curve shape features computed from the training time series, listed in Table~\ref{tab:kg_features}.

\begin{table*}[h]
\centering
\caption{Curve shape features stored per KG node.}
\label{tab:kg_features}
\begin{tabular}{lll}
\toprule
\textbf{Feature} & \textbf{Formula} & \textbf{What it captures} \\
\midrule
Mean             & $\bar{y}$                                      & Average yield level \\
Std deviation    & $\sigma_y$                                     & Spread of yield \\
CV               & $\sigma_y / \bar{y}$                           & Relative volatility \\
Volatility       & $\sigma(\Delta y_t)$                           & Week-on-week smoothness \\
Peak position    & $i^* / N$                                      & Where in season peak occurs (0--1) \\
Zero fraction    & $1 - \frac{1}{N}\sum \mathbf{1}[y_t > 0]$     & Proportion of off-season weeks \\
Early mean       & $\bar{y}_{[0, N/3]}$                           & Average yield in first third \\
Mid mean         & $\bar{y}_{[N/3, 2N/3]}$                        & Average yield in middle third \\
Late mean        & $\bar{y}_{[2N/3, N]}$                          & Average yield in final third \\
Growth pattern   & see below                                      & Seasonal shape classification \\
Harvest window   & start, peak, end ISO week                      & Calendar timing of the season \\
\bottomrule
\end{tabular}
\end{table*}

The growth pattern is assigned by comparing third-by-third seasonal means: \texttt{peak\_middle} if the mid mean exceeds both early and late by more than 10\%; \texttt{increasing} if the late mean exceeds the early mean by more than 10\%; \texttt{decreasing} if the late mean falls below 90\% of the early mean; and \texttt{flat} otherwise.

At prediction time, these features serve three distinct purposes. \textit{Similarity Search} queries the KG for training plots whose curve geometry most closely mirrors the current test entity, providing the agent with analogically grounded historical comparisons. \textit{Dataset Profiling} reads \mbox{\texttt{zero\_fraction}} and \mbox{\texttt{growth\_pattern}} across all nodes to classify the dataset as seasonal or continuous, configuring the agent's phase logic accordingly. \textit{Phase Detection} draws on the stored harvest window to provide the lookahead context required for false-start detection.

Test entities are kept out of the KG entirely to prevent data leakage. Their metadata is extracted separately and passed to the agent prompt only, never used for similarity retrieval.

\subsection{Training Plot Selection}
A grower carries season-to-season memory of how their plots typically behave, drawing on the years that most closely resemble the current one rather than treating all historical records as equally informative. The agent replicates this by ranking each training plot by its average Dynamic Time Warping (DTW) similarity against all test plots, and retaining the top K=50 most representative plots as the training context for the base model. As pairwise DTW comparisons are computationally expensive, this ranking is computed once and cached for reuse on subsequent runs.

\subsection{Agent Architecture and Orchestration}
A human expert does not correct a forecast in a single glance, they gather evidence, form a judgement, act on it, and then reassess. The agent mirrors this deliberative process through a ReAct loop of up to ten iterations per prediction week. Each iteration consists of a single LLM call followed by tool execution. The agent receives a structured prompt containing the current prediction, recent actuals, tool results accumulated so far, and any memory context carried forward from previous plots. It responds with a REASON block explaining its decision and a TOOLS block listing which tools to invoke next. The loop then parses this response, executes the requested tools, updates the agent state, and issues a fresh LLM call with the updated context. The loop exits early when the LLM returns no further tool calls, indicating the agent is satisfied with the correction. If ten iterations are reached without convergence, the last computed correction is accepted as the final output.
In practice, most predictions resolve in two or three iterations: a diagnostic pass to detect the seasonal phase, learn positional bias, and validate the predicted range against historical bounds, followed by a correction pass to apply and verify the result. The ten-iteration ceiling exists as a safeguard against pathological cases rather than as an expected operating regime.

\subsection{Prediction Loop}
\label{sec:prediction_loop}
\subsubsection{Dataset Profiling}
A human grower understands the nature of their crop and operation intuitively, but the agent is dataset-agnostic and must first characterise the data it is working with. Before any predictions are made, the agent inspects the knowledge graph to classify the dataset and configure itself accordingly. For each training entity, the fraction of zero-yield weeks is computed per year. Three entities are sampled and their per-year zero fractions and week-of-year profiles are passed to the LLM, which classifies the dataset as either zero\_valley, where the series genuinely returns to zero during an extended off-season, or positive\_floor, where yields remain above zero year-round. Where no LLM is available, a heuristic fallback applies: if the median per-year zero fraction across sampled entities exceeds 15\%, the dataset is classified as zero\_valley, and positive\_floor otherwise.

\subsubsection{Week-by-Week Prediction Loop}
A grower does not assess all plots simultaneously before moving to the next week; they walk the crop round by round, building a picture of the season incrementally. The prediction loop mirrors this practice by advancing all test plots in lockstep, one week at a time, rather than completing a full plot before moving on. After each plot is processed, its partial curve is committed to the agent's plot history, so find\_similar draws from a progressively richer retrieval pool as the run continues. The position bias table is similarly updated each week using the confirmed error from two weeks prior, once that lagged actual becomes available.

For each week, the system iterates over all test plots and generates a prediction for each before advancing. Each prediction uses the four most recently observed yields ending two weeks before the target date as its context window. A second prediction one week further ahead is generated in parallel, providing the agent with a view of the model's immediate expectation and enabling suppression of isolated noise spikes to zero where appropriate (see false\_start, Section~\ref{sec:phase}). Agent correction is withheld until at least one lagged actual has been recorded; correcting against an empty bias table would be uninformative and risks introducing noise before any systematic error pattern has emerged.

\subsubsection{Agent Prompting}
At each reasoning step, the agent receives a structured prompt containing the raw prediction, available tools with a record of which have already been called, accumulated position bias figures, and a compact set of ordering rules. Replies are constrained to a one-sentence justification and a single tool call, keeping each decision auditable and unambiguous. Any meta-strategy directives synthesised from prior plots are appended at the foot of the prompt (see Section~\ref{para:meta_strategy}).

\subsection{The Agent's Tool Library}
The agent orchestrates a suite of expert-informed tools that function not as isolated scripts, but as a cohesive, sequential diagnostics pipeline. These tools represent holistic heuristics of the agricultural practice, providing the agent with mechanisms to diagnose, bound, and correct predictions statistically.

\subsubsection{Retrieval \& Contextualization: Similarity Search}
Human experts naturally evaluate current crop performance by recalling historically similar seasons, whereas standard ML inference lacks this analogical reasoning. To replicate this, the \textit{Similarity Search} retrieves three historical plots that mathematically mirror the current curve geometry of the target plot. Each plot is encoded as a feature vector, $\mathbf{V}$, encompassing the relative volatility (CV), peak/trough positions ($t^{\text{peak}}, t^{\text{trough}}$), number of peaks \\($\frac{\min(n^{\text{peaks}}, 5)}{5}$ capped at 5), and relative ``spikiness'' to the seasonal mean ($y^{\text{peak}}/\bar{y}$).

\begin{equation}
    \mathbf{v} = \left[\text{CV},\ t^{\text{peak}},\ t^{\text{trough}},\
    \frac{\min(n^{\text{peaks}}, 5)}{5},\ \frac{y^{\text{peak}}}{\bar{y}},\
    \frac{\bar{y}}{y^{\text{peak}}}\right]
\end{equation}

By minimising the Euclidean distance across these vectors  $d(i,j) = \|\mathbf{v}_i - \mathbf{v}_j\|_2$, the agent is able to retrieve plots that share an underlying seasonal distribution, to identify proven historical yield trajectories for further analysis, rather than solely relying on localised raw yields.  

\subsubsection{Diagnostic Evaluation: Bias Learning \& Phase Detection}
These tools assess the baseline prediction in parallel to detect both statistical drift and seasonal phase violations.

\textbf{Bias Learning.}
Statistical forecasts frequently drift into system-wide over or under estimates as the season progresses. In practice, the grower will notice the directional biases and apply a discount or premium to the models future outputs to account for this drift. The \textit{Bias Learning} tool formalises this intuition by evaluating recent predictions against observable lagged actual values. To prevent aggressive corrections base on localised noise, a directional bias is only applied if the ratio of mean error to mean absolute error exhibits a consistent trend ( $\mu_{e} > 0.3 \times \mu_{|e|}$). 

Once confirmed, the magnitude is estimated using the median ratio of actuals to predictions, increasing the robustness to outliers. 
\begin{equation}
    \gamma = \text{median}\left( \frac{y_{\text{actual}}}{y_{\text{predicted}}}
    \right) \quad \text{for } y > 0
\end{equation}.
To prevent hallucinated positive yields post-harvest, this correction factor is exponentially decayed towards neutral as trailing zero-yield weeks accumulate, where $k$ is the number of consecutive trailing zero weeks.
\begin{equation}
    \gamma' = 1 + (\gamma - 1) \cdot 0.5^k \quad \text{where} k > 0,\ \gamma > 1
\end{equation}


\textbf{Phase Detection}
\label{sec:phase}
An experienced grower would immediately identify a baseline model prediction as implausible if it contradicts the current phase of the season; a positive yield prediction before onset, or a rising forecast during a confirmed decline, is immediately suspect. Phase Detection formalises this intuition by segmenting the yield curve into five operational phases derived from curve geometry: pre\_season and ended at the zero-yield boundaries, ramping and declining on the rising and falling slopes, and peak near the seasonal maximum. The tool receives the four most recently observed yields, the raw prediction $\hat{y}$, the forecast horizon $h$, and a one-step-ahead prediction from a second model run. From this, it derives a phase-conditioned estimate $\hat{y}_{\text{phase}}$ and a confidence score $c \in [0,1]$, both forwarded to Apply Correction where $c$ controls how strongly $\hat{y}_{\text{phase}}$ overrides the raw prediction.

\textbf{Seasonal Datasets.}
Phases are evaluated in priority order and the first matching condition is returned. pre\_season, where no non-zero yield exists, sets $\hat{y}_{\text{phase}} = 0$ with $c = 0.90$; a false\_start sub-case is detected when the model predicts a positive yield but the one-step-ahead prediction is near-zero, indicating an isolated noise spike before onset ($c = 0.85$). ended, where one or more trailing zero weeks follow the last non-zero week $y_l$, also sets $\hat{y}_{\text{phase}} = 0$, with confidence scaling as $c = \min(0.95,\ 0.75 + 0.05k)$ where $k$ is the number of trailing zero weeks. The early phase, where only a single non-zero week has been observed, sets $\hat{y}_{\text{phase}} = \max(\hat{y},\ y_l)$ with low confidence ($c = 0.35$) to avoid undercutting the sole observation. During declining, a per-week decay rate (Eq.~\ref{eq:decay_rate}) is fitted to post-peak observations and projected forward by horizon $h$ (Eq.~\ref{eq:bio_estimate}):
\begin{equation}
    \delta = \left(\frac{y_l}{y_{\text{peak}}}\right)^{1/\max(\Delta t,\, 1)}
    \label{eq:decay_rate}
\end{equation}
\begin{equation}
    \hat{y}_{\text{phase}} = y_l \cdot \delta^{\,h}
    \label{eq:bio_estimate}
\end{equation}
with values below 1.0 floored to zero, and confidence growing with the active season window $n_a$ (the span from first to last non-zero week, including any gaps) as given in Eq.~\ref{eq:declining_confidence}:
\begin{equation}
    c = \min(0.85,\ 0.4 + 0.1 \cdot n_a)
    \label{eq:declining_confidence}
\end{equation}
Active growth phases (ramping, peak) defer to the model with $\hat{y}_{\text{phase}} = \hat{y}$, as the yield trajectory remains least predictable, with peak carrying a moderate confidence of $c = 0.40$ and ramping fully trusting the model at $c = 0.0$.

\textbf{Continuous Datasets.}
On continuous datasets, where no seasonal bell curve exists, the tool instead 
splits the recent window in half and compares the means of each half:
\begin{equation}
    \text{change} = \frac{\bar{y}_{\text{second}} - \bar{y}_{\text{first}}}{|\bar{y}_{\text{first}}|}
    \label{eq:continuous_change}
\end{equation}
A change exceeding $+10\%$ is classified as trending\_up 
(Eq.~\ref{eq:continuous_change}), below $-10\%$ as trending\_down, and 
otherwise stable. In all three cases $\hat{y}_{\text{phase}} = \hat{y}$. 
Confidence is capped at $0.75$ and scales with the magnitude of the change, 
so a weak trend carries little weight.

\subsubsection{Empirical Bounding: Range Validation}

Domain experts possess an intuitive understanding of plausible weekly growth rates, and will immediately reject a forecast that suggests an unprecedented leap that defies the known crop physiology. Therefore, \textit{Range Validation} acts as a pre-correction filter, computing the fractional change from the last observed yield to the new prediction. The change is evaluated against the 10th and 90th percentiles of the historical changes observed at the exact same seasonal position. 
\begin{equation}
    \Delta\% = \frac{\hat{y} - y_{t-1}}{y_{t-1}} \in [P_{10} - 0.5,\ P_{90} + 0.5]
\end{equation}
Predictions breaching these empirical boundaries are clamped to the nearest plausible value, albeit with a 50\% buffer outside the historical range. If a predicted change exceeds these limits, it is mathematically clamped to $\hat{y}=y_{t-1}\cdot(1+P_{90})$ for an upward breach, or $\hat{y}=y_{t-1}\cdot(1+P_{10})$ for a downward breach. However, if fewer than three historical comparisons are available at the current position, the tool defers and returns in-range by default. This ensures the proposed change is historically plausible before any further synthesis occurs.

\subsubsection{Synthesis: Apply Correction}
An expert must weigh conflicting pieces of evidence, prioritising physical constraints over statistical patterns. This tool integrates the empirical bounds and diagnostic outputs described above to resolve the final corrected value ($\hat{y}^*$) through a strict three-tiered priority cascade, stopping as soon as a condition is met to prevent cascading errors. At the highest priority is the \textit{Physical Limit Check}: if an implausible leap is flagged, the value is clamped to the nearest historical boundary, superseding all other logic. If the physical limits are respected, the system moves to the \textit{Phase Check}. Here, if phase detection confidence is high ($c \geq 0.5$), the raw prediction is blended with the phase-conditioned estimate using the formula $\hat{y}^* = c \cdot \hat{y}_{\text{phase}} + (1 - c) \cdot \hat{y}_{\text{raw}}$. When this phase blend is triggered, statistical multipliers are intentionally suppressed to avoid double-correction. Finally, if neither physical nor phase constraints are triggered, the system defaults to \textit{Statistical Fine-Tuning}, applying the position bias correction and learned bias multiplier to adjust the final output.

\subsubsection{Verification: Safety Verifier \& Trajectory Evaluation}

Even human expert interventions require sanity check to ensure adjustments do not result in catastrophic algorithmic drift. The \textit{Safety Verifier} prevents agent hallucinations by analysing the final correction ratio ($R = \hat{y}^* / \hat{y}_{\text{raw}}$) alongside absolute limits. Results that fall in the warning zone—where $R \in [P_{01}, P_{05}]$ or $[P_{95}, P_{99}]$ are flagged in the output logs but not overridden, allowing unusual but historically possible corrections to pass through. However, if the ratio exceeds the extreme 1st or 99th percentiles of the training distribution, predicts positive yields during dormant off-seasons, or exceeds three times the historical maximum the verifier forcibly overrides the agent and clamps the output to predefined safe boundaries. This boundary is defined as the \textit{historical maximum clamp} where if $\hat{y}^* > 3 \times \text{hist\_max}$ (where $\text{hist\_max}$ is the highest value seen across all training plots), the corrected value is immediately clamped to $\text{hist\_max}$.

Simultaneously, \textit{Trajectory Evaluation} acts as a self-critique mechanism to ensure the corrected value aligns with recent trends and the expected seasonal phase. It projects a plausible range for the upcoming week based on the last three observed yields, calculating a 30\% tolerance above and below the trend-projected value: $[\hat{y}_{\text{lo}}, \hat{y}_{\text{hi}}] = \bigl[0.7 \cdot \min(y_t, y_t(1+\delta)), 1.3 \cdot \max(y_t, y_t(1+\delta))\bigr]$, where the fractional trend slope $\delta = (y_t - y_{t-2}) / (2 \cdot \bar{w})$ is normalised by the three-week mean $\bar{w}$. 

The tool evaluates the correction against this range to issue a verdict, returning ``consistent'' if the value is in-range and phase-sound; ``trend contradiction'' if it aligns with the current phase but exceeds the mathematical range; and ``phase contradiction'' if it opposes the current seasonal phase, such as a downward correction during a confirmed ramping phase. If a contradiction in either trend or phase is detected, it triggers an \textit{Iterative Adjustment}. This prompts the agent to dynamically recalibrate by choosing a revised blend weight ($w'$), allowing it to soften an overcorrection or adjust an undercorrection based on the specific failure mode. The agent then re-executes the correction pipeline using the updated formula: $\hat{y}^* \leftarrow w' \cdot \hat{y}_{\text{phase}} + (1 - w') \cdot \hat{y}_{\text{raw}}$. To ensure stability and prevent runaway loops, this iterative adjustment is strictly capped at a maximum of two attempts per prediction week.

\subsection{Cross-Plot Persistent Memory}
Up to this point, the correction pipeline has functioned as a reactive system: detecting phases, enforcing physical limits, and smoothing trajectories based purely on the immediate context of a single plot. However, if the agent resets its context after every plot, it remains blind to broader systematic errors and risks repeating the same correction mistakes across the dataset. To solve this, the pipeline is augmented with a long-term memory mechanism.

First, a \textit{Position Bias Table} continuously records percentage prediction errors mapped to specific spatial-temporal coordinates (week-of-year and season-progress). This allows the system to learn and proactively correct when the base model systematically fails at specific calendar milestones. Second, the \textit{Plot History} archives the geometric shapes and final correction outcomes of every completed plot, continually expanding the dynamic retrieval corpus for the Similarity Search tool. 

Finally, a \textit{Meta-Strategy Reflection} protocol \label{para:meta_strategy} occurs at regular intervals. The agent processes a summary of recent correction outcomes, assessing tool efficacy and impact on mean absolute error. It synthesizes these findings into abstract strategic directives such as noting the counter-productivity of a specific tool during early growth, which are permanently injected into subsequent prompts. This enables the agent to autonomously refine its experiential heuristics over the lifespan of the dataset run.


\subsection{Experimental Setup}
Three LLMs were evaluated for agent reasoning: \texttt{llama3.1:8b}, \texttt{qwen2.5:7b}, are text-based instruction-following models suited to structured numerical reasoning tasks. \texttt{Llava:13b} is a vision-language model included to examine the sensitivity of the framework to the choice of LLM model for refinement. Its pretraining is oriented toward image understanding rather than structured numerical reasoning, which makes it mismatched to this task. Where LLaVA produces inconsistent corrections and often degrades performance, this highlights that the choice of LLM model for refinement is a meaningful design decision. The agent framework provides the structure, but the underlying LLM must be capable of reliably following instructions with numerical context to be effective. All three models were hosted locally via Ollama with no external API calls. Baselines are XGBoost, Random Forest, and Moirai2 without agent correction.

\subsection{Forecasting Models}
Three base forecasting models are evaluated as the prediction backbone, spanning classic gradient-boosted approaches and a modern zero-shot foundation model.

\textbf{XGBoost} is a gradient-boosted tree model trained on windowed yield and weather features. Despite lacking inherent temporal structure, it has proven competitive in yield forecasting tasks~\cite{beddows2024multi} and outperforms specialised time series models such as ARIMA in similar
settings~\cite{alim2020comparison}. Three separate models are trained, one per quantile ($q_{10}$, $q_{50}$, $q_{90}$), using quantile loss to produce a lower bound, median point forecast, and upper bound for each prediction. This gives the agent a direct signal of model uncertainty, and provides actionable prediction intervals for operational planning, since a grower needs to know the range of plausible outcomes rather than a single figure.

\textbf{Random Forest} is an ensemble method that aggregates predictions across multiple decision trees, and has been widely applied to crop yield forecasting~\cite{prasad2021crop}. It is trained on the same windowed yield and weather features as XGBoost, producing a single point forecast per prediction step.

\textbf{Moirai 2}~\cite{liu2025moirai} is a zero-shot time series foundation model pre-trained on 36 million series comprising approximately 295 billion observations. Unlike its predecessor \cite{woo2024moirai}, it adopts a decoder-only autoregressive architecture with quantile-based predictions and multi-token prediction, improving both probabilistic accuracy and inference efficiency.

\section{Results}

\subsection{Overall Performance}

\begin{table*}[]
\centering
\caption{Performance comparison of various baseline models refined by LLMs (Normalized).}
\label{tab:multi_model_comparison}
\begin{tabular}{llcccccc}
\toprule
& & \multicolumn{3}{c}{\textbf{Angus Strawberry}} & \multicolumn{3}{c}{\textbf{USDA Corn}} \\
\cmidrule(lr){3-5} \cmidrule(lr){6-8}
\textbf{Baseline} & \textbf{Refinement Model} & \textbf{MAE} & \textbf{RMSE} & \textbf{MASE} & \textbf{MAE} & \textbf{RMSE} & \textbf{MASE} \\
\midrule
\multirow{5}{*}{\textbf{XGBoost}}
    & None (Baseline) & 0.0131 & 0.0433 & 3.9313 & 0.0244 & 0.0721 & 0.9964 \\
    & Llama 3.1 8B    & \textbf{0.0105} & \textbf{0.0399} & \textbf{1.7454} & \textbf{0.0235} & 0.0714 & \textbf{0.9716} \\
    & LLaVA 13B       & 0.0124 & 0.0415 & 3.7192 & 0.0257  & 0.0801 & 1.0489      \\
    & Qwen 2.5 7B     & 0.0111 & 0.0404 & 2.0965 & 0.0239 & \textbf{0.0708} & 0.9887 \\
\midrule
\multirow{5}{*}{\textbf{Moirai2}}
    & None (Baseline) & 0.0161 & 0.0519 & 1.7834 & 0.0500 & 0.1182 & 2.1705 \\
    & Llama 3.1 8B    & \textbf{0.0123} & \textbf{0.0446} & \textbf{1.3868} & \textbf{0.0379} & \textbf{0.0990} & \textbf{1.6290} \\
    & LLaVA 13B       & 0.0152 & 0.0500 & 1.6890 & 0.0487 & 0.1178 & 2.1146 \\
    & Qwen 2.5 7B     & 0.0140 & 0.0481 & 1.5707  & 0.0427 & 0.1067 & 1.8395 \\
\midrule
\multirow{5}{*}{\textbf{Random Forest}}
    & None (Baseline) & 0.0153 & 0.0410 & 5.9036 & 0.0265 & \textbf{0.0646} & 1.0843 \\
    & Llama 3.1 8B    & \textbf{0.0110} & \textbf{0.0365} & \textbf{2.0304} & \textbf{0.0236} & 0.0654 & \textbf{0.9711} \\
    & LLaVA 13B       & 0.0147 & 0.0404 & 5.5686 & 0.0292 & 0.0765 & 1.2049 \\
    & Qwen 2.5 7B     & 0.0132 & 0.0381 & 5.1964 & 0.0259 & 0.0661 & 1.0648 \\
\bottomrule
\end{tabular}
\end{table*}

Table~\ref{tab:multi_model_comparison} reports normalised MAE, RMSE, and MASE across all evaluated configurations

\subsection{Effect of Agent Refinement on XGBoost}
Across both datasets, agent-based refinement consistently improved on the unrefined 
XGBoost baseline. On the Angus Strawberry dataset, Llama 3.1 8B achieved the largest gains, reducing MAE from 0.0131 to 0.0105 ($-$20\%) and MASE from 3.93 to 1.75 ($-$56\%). Qwen 2.5 7B followed closely, reducing MAE to 0.0111 and MASE to 2.10. LLaVA 13B produced more modest improvements in MAE (0.0131 $\to$ 0.0124) and RMSE, but MASE remained near the baseline (3.72 vs.\ 3.93), indicating that while absolute errors were reduced, the forecast did not improve substantially compared to the base model.

On USDA Corn, all three LLMs reduced the MAE and RMSE, though increases in accuracy  were much smaller, consistent with the lower baseline error on this dataset. Qwen 2.5 7B achieved the best RMSE (0.0708, $-$1.8\%), while Llama 3.1 8B achieved the best MASE (0.9716, $-$2.5\% over the baseline of 0.9964). LLaVA 13B was the only configuration to increase error on corn (MAE $+$5.3\%, RMSE $+$11.1\%), as expected due to its vision-language pretraining providing limited benefit for numeric timeseries correction.

\subsection{Effect of Agent Refinement on Moirai2}
The Moirai2 foundation model provided a stronger strawberry baseline than XGBoost 
in MASE terms (1.78 vs.\ 3.93), but a weaker corn baseline (MAE 0.0500 vs.\ 
0.0244). Agent refinement improved results across all three LLMs on both datasets.

On strawberry, Llama 3.1 8B reduced MAE by 23.6\% (0.0161 $\to$ 0.0123) and 
MASE by 22.2\% (1.78 $\to$ 1.39). Qwen 2.5 7B achieved a 13.1\% MAE reduction 
and brought MASE to 1.57. LLaVA 13B produced smaller but consistent gains, reducing the MASE from 1.78 to 1.69.

On corn, the Moirai2 baseline was much weaker, agent corrections were most impactful. Llama 3.1 8B reduced the MAE by 24.2\% (0.0500 $\to$ 0.0379) and the MASE by 24.9\% (2.17 $\to$ 1.63). Qwen 2.5 7B achieved a 14.6\% MAE reduction (MASE 1.84). LLaVA 13B again produced the smallest improvement, reducing the MASE only marginally (2.17 $\to$ 2.11). Figure~\ref{fig:rf_llama} illustrates the correction pattern for Random Forest + Llama 3.1 8B across both datasets.

\subsection{Effect of Agent Refinement on Random Forest}
The Random Forest baseline has a high MASE on strawberry (5.90) relative to XGBoost and Moirai2, reflecting the mean RF's tendency to predict small non-zero yields during the off-season. This provides the agent with a clear correction signal, and Llama 3.1 8B produced the strongest correction of any configuration in the study: MAE $-$28.1\% (0.0153 $\to$ 0.0110) and MASE $-$65.6\% (5.90 $\to$ 2.03). Qwen 2.5 7B achieved a 13.7\% MAE reduction (MASE 5.20), while LLaVA 13B again contributed only marginal improvement (MAE $-$3.9\%, MASE 5.57).

On the corn dataset, the RF baseline performed better (MAE 0.0265, MASE 1.08), Llama 3.1 8B reduced the MAE by 11.0\% and brought the MASE below 1.0 (0.97). Qwen 2.5 7B produced modest gains ($-$2.3\% MAE) and LLaVA 13B degraded the performance on corn ($+$10.2\% MAE, $+$18.4\% RMSE), consistent with its pattern of unreliable corrections on the structured corn signal. Figure~\ref{fig:rf_llama} illustrates the correction pattern for Random Forest + Llama 3.1 8B across both datasets.

\begin{figure*}[h]
    \centering
    \includegraphics[width=1.0\textwidth]{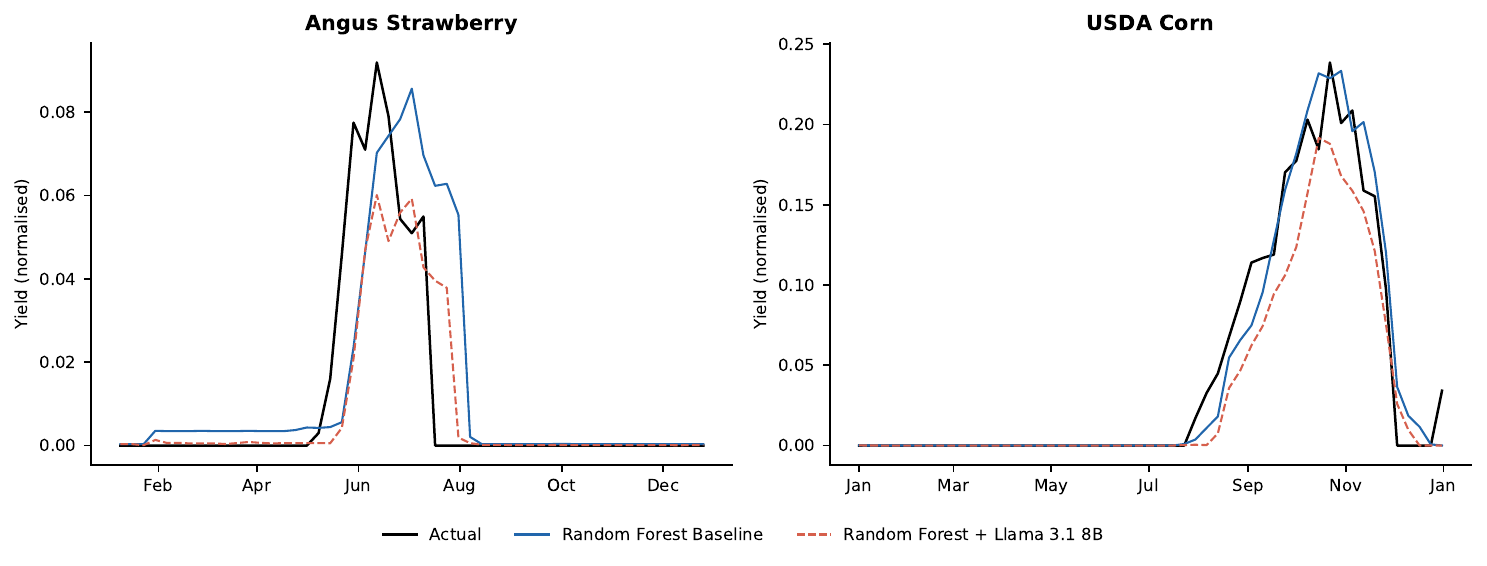}
    \caption{Random Forest Llama 3.1 on both datasets.}
    \label{fig:rf_llama}
\end{figure*}

\subsection{Cross-Model Analysis}
Llama 3.1 8B was the most effective refinement model in every configuration, achieving the best MAE and MASE in all twelve baseline/dataset combinations. Qwen 2.5 7B scored second, producing moderate improvements. LLaVA 13B was the weakest and most inconsistent model, often degrading results on corn for both XGBoost and Random Forest. Figure~\ref{fig:scatter} shows the per-plot distribution of corrections for XGBoost + Llama 3.1 8B, with the majority of plots falling below the diagonal on both datasets.

\begin{figure*}[]
\centering
\includegraphics[width=0.85\textwidth]{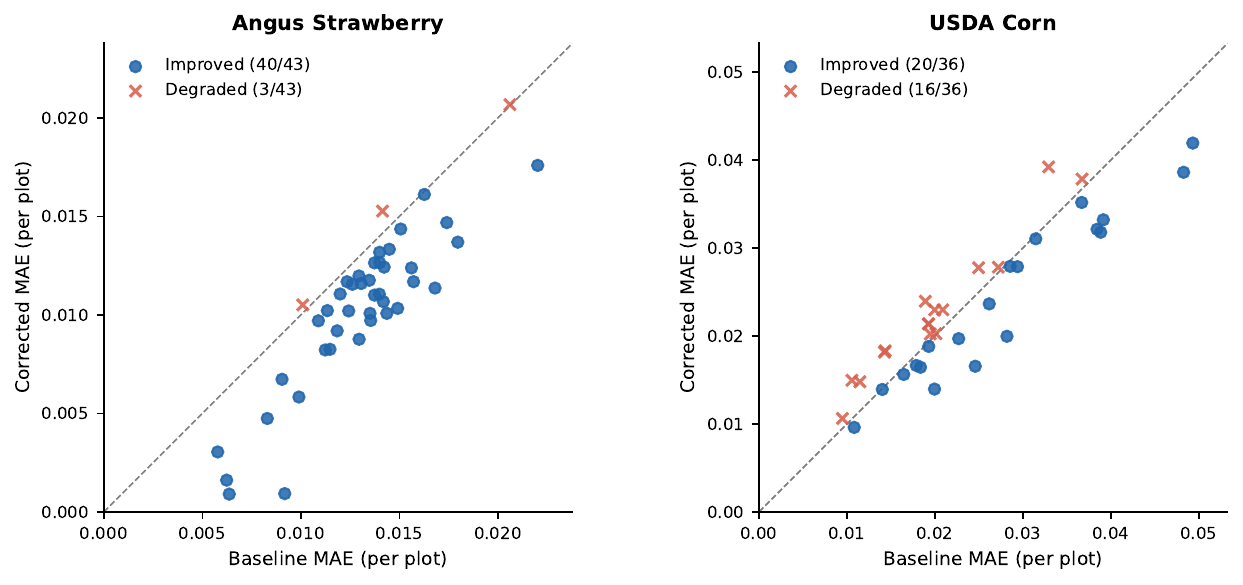}
\caption{Per-plot MAE before and after agent correction (XGBoost + Llama 3.1 8B). 
Points below the diagonal indicate improvement. Blue circles = improved plots; 
red crosses = degraded plots.}
\label{fig:scatter}
\end{figure*}

\subsection{Agent Tool Utilisation}

On average, the agent invokes 6.4 tools per prediction. Table~\ref{tab:tools} breaks this down by tool and phase. The analysis phase accounts for the bulk of calls: \texttt{learn\_bias} and \texttt{detect\_phase} are invoked on virtually every prediction (99.6\%), providing the bias estimate and growth-phase context that anchor the correction. \texttt{find\_similar} (84.1\%) and \texttt{validate\_range} (73.0\%) are called conditionally, the agent skips them when historical analogues are sparse or the proposed correction falls within expected bounds without further checks.

Correction application (\texttt{apply\_correction}, 98.7\%) is near-universal, followed by \texttt{evaluate\_trajectory} (67.8\%), which assesses whether the corrected value is consistent with the seasonal trend to date. When the trajectory assessment is unfavourable, the agent invokes \texttt{adjust\_correction} to refine the correction (19.7\% of all predictions, or roughly one in three trajectory evaluations). \texttt{verify\_correction} closes the loop on 98.7\% of predictions, confirming the final value is physically plausible before it is accepted.

\begin{table}[t]
\centering
\caption{Agent tool invocation frequency across all non-trivial predictions on the strawberry dataset (XGBoost + Llama~3.1~8B, $n=233$).}
\label{tab:tools}
\begin{tabular}{llr}
\toprule
\textbf{Phase} & \textbf{Tool} & \textbf{Call Rate} \\
\midrule
\multirow{4}{*}{Analysis}
 & \texttt{learn\_bias}          & 99.6\% \\
 & \texttt{detect\_phase}        & 99.6\% \\
 & \texttt{find\_similar}        & 84.1\% \\
 & \texttt{validate\_range}      & 73.0\% \\
\midrule
\multirow{2}{*}{Correction}
 & \texttt{apply\_correction}    & 98.7\% \\
 & \texttt{adjust\_correction}   & 19.7\% \\
\midrule
\multirow{2}{*}{Verification}
 & \texttt{evaluate\_trajectory} & 67.8\% \\
 & \texttt{verify\_correction}   & 98.7\% \\
\bottomrule
\end{tabular}
\end{table}

\subsection{Ablation Study}
\label{sec:ablation}

To quantify the contribution of each agent tool, we conduct a leave-one-out ablation on the best-performing configuration (XGBoost baseline, Llama 3.1 8B refinement) across both datasets. In each combination a single tool is disabled while the remainder of the pipeline operates normally. \texttt{evaluate\_trajectory} and \texttt{adjust\_correction} are disabled together, as the latter has no meaning without the former.Table~\ref{tab:ablation} reports the corrected MAE, RMSE, and MASE for each condition. Table~\ref{tab:ablation_reverse} reports the reverse ablation, where each tool is enabled in isolation, confirming that \texttt{detect\_phase} alone accounts for the majority of the pipeline's correction capability.

\begin{table*}[]
\centering
\caption{Ablation study: effect of removing individual agent tools. 
XGBoost baseline, Llama 3.1 8B, leave-one-out per condition.}
\label{tab:ablation}
\begin{tabular}{lcccccc}
\toprule
& \multicolumn{3}{c}{\textbf{Angus Strawberry}} & \multicolumn{3}{c}{\textbf{USDA Corn}} \\
\cmidrule(lr){2-4} \cmidrule(lr){5-7}
\textbf{Configuration} & \textbf{MAE} & \textbf{RMSE} & \textbf{MASE} & \textbf{MAE} & \textbf{RMSE} & \textbf{MASE} \\
\midrule
Full system               & 0.0105 & 0.0399 & 1.7454 & 0.0235 & 0.0714 & 0.9716 \\
\midrule
$-$\texttt{learn\_bias}          & 0.0101 & 0.0388 & 1.6863 & 0.0246 & 0.0746 & 1.0112 \\
$-$\texttt{detect\_phase}        & 0.0222 & 0.0895 & 4.9232 & 0.0356 & 0.1261 & 1.4997 \\
$-$\texttt{validate\_range}      & 0.0105 & 0.0392 & 1.7507 & 0.0240 & 0.0744 & 0.9846 \\
$-$\texttt{find\_similar}        & 0.0103 & 0.0390 & 1.6649 & 0.0236 & 0.0721 & 0.9682 \\

$-$\texttt{evaluate\_trajectory} & 0.0100 & 0.0387 & 1.6488 & 0.0252 & 0.0750 & 1.0504 \\
$-$\texttt{verify\_correction}   & 0.0109 & 0.0389 & 2.3435 & 0.0242 & 0.0727 & 0.9989 \\

\midrule
None (XGBoost baseline)   & 0.0131 & 0.0433 & 3.9313 & 0.0244 & 0.0721 & 0.9964 \\
\bottomrule
\end{tabular}
\end{table*}

\begin{table*}[h]
\centering
\caption{Reverse ablation study: effect of enabling a single agent tool in isolation. 
XGBoost baseline, Llama 3.1 8B.}
\label{tab:ablation_reverse}
\begin{tabular}{lcccccc}
\toprule
& \multicolumn{3}{c}{\textbf{Angus Strawberry}} & \multicolumn{3}{c}{\textbf{USDA Corn}} \\
\cmidrule(lr){2-4} \cmidrule(lr){5-7}
\textbf{Configuration} & \textbf{MAE} & \textbf{RMSE} & \textbf{MASE} & \textbf{MAE} & \textbf{RMSE} & \textbf{MASE} \\
\midrule
Full system               & 0.0105 & 0.0399 & 1.7454 & 0.0235 & 0.0714 & 0.9716 \\
\midrule
\texttt{learn\_bias} only          & 0.0231 & 0.0891 & 4.9522 & 0.0377 & 0.1335 & 1.6057 \\
\texttt{detect\_phase} only        & 0.0101 & 0.0389 & 2.0102 & 0.0275 & 0.0805 & 1.1584 \\
\texttt{validate\_range} only      & 0.0125 & 0.0413 & 3.7479 & 0.0240 & 0.0701 & 0.9774 \\
\texttt{find\_similar} only        & 0.0129 & 0.0421 & 3.8310 & 0.0244 & 0.0721 & 0.9944 \\
\texttt{evaluate\_trajectory} only & 0.0129 & 0.0421 & 3.8274 &  0.0244  & 0.0721 & 0.9964  \\
\texttt{verify\_correction} only   & 0.0129 & 0.0421 & 3.8273 & 0.0241 & 0.0716 & 0.9992 \\
\midrule
None (XGBoost baseline)   & 0.0131 & 0.0433 & 3.9313 & 0.0244 & 0.0721 & 0.9964 \\
\bottomrule
\end{tabular}
\end{table*}

Removing \texttt{detect\_phase} causes the largest degradation across both datasets, MAE increases by 72.0\% on strawberry (0.0105 $\to$ 0.0222) and 45.7\% on corn (0.0235 $\to$ 0.0356) compared to the full system, with RMSE more than doubling on strawberry (0.0399 $\to$ 0.0895). This confirms that phase detection is the load-bearing component of the pipeline: without it, the agent lacks the context needed to reason about whether a prediction error stems from a correctable trend or a local deviation, causing corrections to become unreliable.

Removing \texttt{verify\_correction} produces a notable MASE increase on strawberry (1.7454 $\to$ 2.3435) with only minor MAE change (+3.8\%). Notably, several ablations marginally outperform the full system on MAE, likely reflecting noise at this scale of improvement rather thangenuine redundancy. Taken together, the ablation suggests the pipeline is robust
to the removal of any single secondary tool, but critically dependent on 
\texttt{detect\_phase} to anchor the correction process.

\subsection{Computational Cost Analysis}
\label{sec:compute}
Without agent correction, the base XGBoost model completes inference in 31.34 seconds on the strawberry dataset (43 plots, 2,193 predictions) and 24.42 seconds on corn (36 plots, 1,908 predictions). Enabling Llama 3.1 8B correction increases wall time to 2.83 hours and 2.44 hours respectively, reflecting the cost of the ReAct loop issuing over 20,000 LLM calls per dataset, averaging 827 prompt tokens and 40 output tokens per call.

All inference was performed locally on an NVIDIA GeForce RTX 3070 (8\,GB). LLM prompts were sent over a local network tunnel to a separate machine with an NVIDIA GeForce RTX 3090 (24\,GB), with the responses returned to the host machine for use in the correction pipeline.

\section{Discussion}
The magnitude of agents prediction improvement correlates with baseline error. Combinations where the baseline model has lower accuracy (e.g., RF on strawberry, Moirai2 on corn) show the largest gains in performance. This is consistent with the agent's role as a post prediction corrector, the tools for \textbf{bias learning} and \textbf{phase detection} are most informative when the baseline prediction diverges substantially from expected phase patterns. Where the baseline prediction is already accurate (XGBoost on corn, MASE $\approx 1.0$), the agent's room for improvement is limited and the corrections it makes are expectedly smaller.

[Discussion of when agents help most, comparison to base models.]

Figure~\ref{fig:improvement_bar} shows the per-plot MAE improvement distribution for XGBoost + Llama 3.1 8B across both datasets. The majority of plots benefit from the agents corrections, but a minority are degraded. Instances of the model degrading the predictions is mostly on the corn dataset, where LLaVA 13B produced the largest individual losses.

\begin{figure*}[h]
    \centering
    \includegraphics[width=1.0\textwidth]{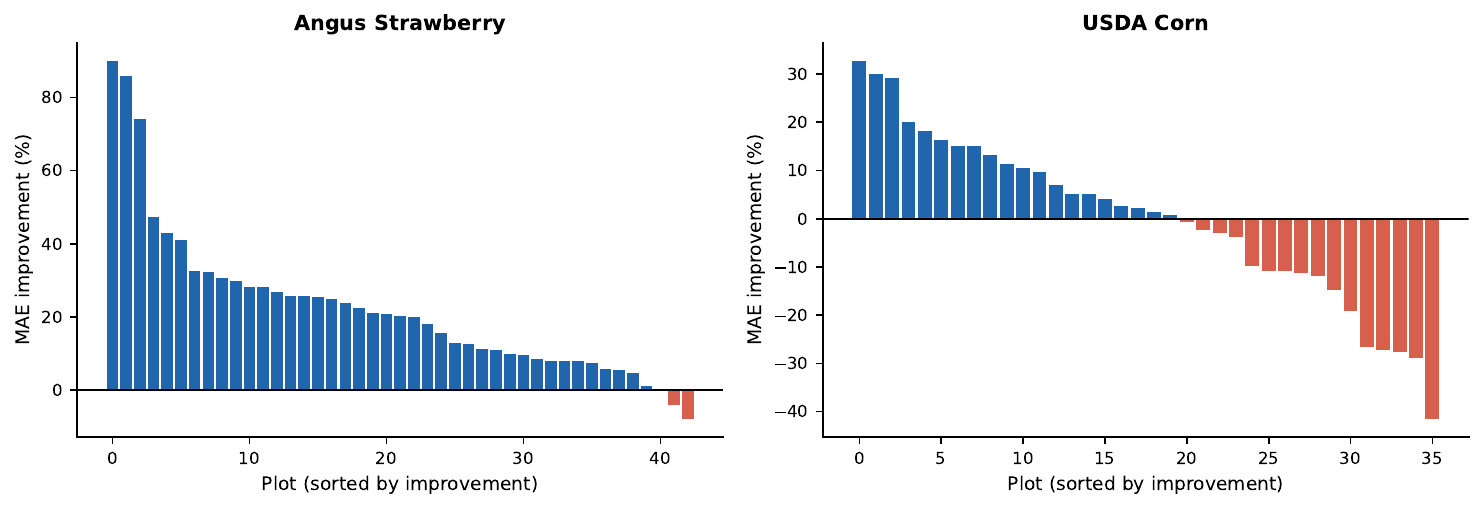}
    \caption{Per-plot MAE improvement (\%) for XGBoost + Llama 3.1 8B, sorted 
    descending. Blue bars indicate improvement; red bars indicate degradation.}
    \label{fig:improvement_bar}
\end{figure*}

\subsection{Generalizability and Broader Applicability}
The agent framework was designed to be dataset-agnostic. The tool set encodes seasonal reasoning through seasonal structure, phase detection, decay estimation, and historical range validation. None of these have been hard coded to match the yields of the dataset we used, or strawberries in general. The USDA corn results support this: the same agent with no modification the to tools or prompting, was able to produce consistent improvements on a structurally similar but agriculturally distinct crop, harvested at a different scale, in a different country.

The system requires at least weekly yield records and a seasonal growth curve with an identifiable peak and post-peak decline. This structure is common across soft fruit and most field crops. Crops with irregular yield patterns or continuous production without a clear off-season would require adaptation to the phase detection logic.

\subsection{Limitations}
\paragraph{Computational Overhead}\mbox{}\\
The agent introduces substantial computational cost relative to the base forecasting model, as detailed in Section~\ref{sec:compute}. This overhead stems from the nature of the ReAct loop, which issues one or more LLM calls per prediction step. All inference was performed locally on a single GPU via Ollama; wall time is therefore hardware-dependent, and token counts provide a more portable measure of computational demand. In a production setting this cost could be reduced by batching calls or using a faster inference backend, 
but as evaluated, the agent is not suitable for real-time or low-latency deployment without further optimisation.

\paragraph{Dependence on LLM Capabilities}\mbox{}\\
The quality of the corrections made by the agent is directly tied to the capabilities of the underlying language model. Smaller models, such as those evaluated in this work, may occasionally produce incorrect tool sequences or fail to appropriately weight the numerical context they are provided with. This means the performance of the system is not solely determined by the reasoning encoded in the toolset, but also by how reliably the model follows structured instructions.This can mean that results are not fully reproducible across different model versions. This is demonstrated by the inferior performance shown when using the LLaVA 13B model as the refinement agent.

\paragraph{Explainability of Agent Decisions}\mbox{}\\
While each individual tool call can be inspected, the overall correction is computed from the LLM synthesising multiple tool outputs, which is not fully transparent. A fixed heuristic correction method is auditable and guaranteed to behave consistently, whereas the agent may reach the same conclusion via different reasoning paths on different runs. In a commercial setting, where a grower may want to understand why a forecast has been adjusted, this is a meaningful limitation.

\section{Conclusion}
In this paper, we present a structured LLM agent framework for post-hoc correction of agricultural yield forecasts, designed for the data-constrained conditions that characterise commercial soft fruit production. Rather than addressing the scarcity of farm data by acquiring additional inputs, the agent reasons more effectively over what is already available, applying crop-grounded and statistical logic as a correction layer on top of an existing forecasting model.

The results demonstrate that this approach consistently improves forecast accuracy across three base models and two structurally similar but agriculturally distinct datasets. The largest gains occur where the baseline model diverges most from expected seasonal behaviour, confirming that the agent's value lies in correcting systematic errors that the base model lacks the context to fix itself. Llama 3.1 8B produced the strongest corrections across all configurations, while LLaVA 13B's inconsistent performance highlights that the choice of refinement model is a meaningful design decision.

The ablation study identifies phase detection as the load-bearing component of the pipeline. Without it, corrections become unreliable across both datasets. The remaining tools contribute incrementally to the overall correction quality.

The framework requires no retraining, no additional data collection, and no modification to the base forecasting model. The corn results were produced with no modification to the agent, tools, or prompting. We believe this positions agent-based post-hoc correction as a practical and deployable complement to existing forecasting pipelines in agricultural yield forecasting under realistic data conditions.

\section*{Data Availability}
The strawberry yield dataset was provided under a commercial agreement and cannot be made publicly available. The corn dataset is publicly available from the USDA NASS QuickStats database at \url{https://quickstats.nass.usda.gov/} and can be reproduced using the queries described in Section~\ref{sec:datasets}.

\section*{Funding}
This work was funded by the University of Aberdeen, The Data Lab, and Angus Soft Fruits Ltd as part of a PhD studentship. 

\section*{Acknowledgments}
We would like to thank Angus Soft Fruits for their continued support and data access.

\section*{Conflicts of Interest}
The authors declare no conflicts of interest.

\bibliographystyle{plain}

\end{document}